\documentclass[10pt,twocolumn,letterpaper]{article}

\usepackage{iccv}
\usepackage{times}
\usepackage{epsfig}
\usepackage{graphicx}
\usepackage{amsmath}
\usepackage{amssymb}

\usepackage{xcolor}
\usepackage{url}            
\usepackage{booktabs}       
\usepackage{amsfonts}       
\usepackage{nicefrac}       
\usepackage{microtype}      
\usepackage{subfig}
\usepackage{multirow}
\usepackage{makecell}
\usepackage{xspace}
\usepackage{mathabx}
\usepackage{mathtools}
\usepackage{stmaryrd}
\usepackage{bm}
\usepackage[square,numbers,sort,compress]{natbib}

\usepackage[pagebackref=true,breaklinks=true,letterpaper=true,colorlinks,bookmarks=false]{hyperref}

\usepackage{cleveref}

\iccvfinalcopy 

\def\iccvPaperID{5029} 

\newcommand\BB{\mathcal{B}}
\newcommand\CC{\mathcal{C}}
\newcommand\EE{\mathcal{E}}
\newcommand\EXP{\mathop{\mathbb{E}}}
\newcommand\stopgrad[1]{\langle #1 \rangle}
\newcommand\Stopgrad[1]{\left\langle #1 \right\rangle}
\newcommand{\vect}[1]{\boldsymbol{#1}}

\newcommand{\norm}[1]{\left\lVert#1\right\rVert}

\newcommand{\cifarten}{CIFAR-10\xspace}
\newcommand{\cifarhun}{CIFAR-100\xspace}

\newcolumntype{L}[1]{>{\raggedright\let\newline\\\arraybackslash\hspace{0pt}}m{#1}}
\newcolumntype{C}[1]{>{\centering\let\newline\\\arraybackslash\hspace{0pt}}m{#1}}
\newcolumntype{R}[1]{>{\raggedleft\let\newline\\\arraybackslash\hspace{0pt}}m{#1}}

\ificcvfinal\fi
\begin{document}

\title{EvalNorm: Estimating Batch Normalization Statistics for Evaluation}

\author{Saurabh Singh\\
Google Research\\
{\tt\small saurabhsingh@google.com}
\and
Abhinav Shrivastava\thanks{Work done while at Google.}\\
University of Maryland, College Park\\
{\tt\small abhinav@cs.umd.edu}
}

\maketitle
\ificcvfinal\thispagestyle{empty}\fi

\begin{abstract}
Batch normalization (BN) has been very effective for deep learning and is widely used. However, when training with small minibatches, models using BN exhibit a significant degradation in performance. In this paper we study this peculiar behavior of BN to gain a better understanding of the problem, and identify a cause. We propose `EvalNorm' to address the issue by estimating corrected normalization statistics to use for BN during evaluation. EvalNorm supports online estimation of the corrected statistics while the model is being trained, and does not affect the training scheme of the model. As a result, EvalNorm can also be used with existing pre-trained models allowing them to benefit from our method. EvalNorm yields large gains for models trained with smaller batches. Our experiments show that EvalNorm performs 6.18\% (absolute) better than vanilla BN for a batchsize of 2 on ImageNet validation set and from 1.5 to 7.0 points (absolute) gain on the COCO object detection benchmark across a variety of setups.
\end{abstract}

\section{Introduction}\label{sec:intro}

\begin{figure*}[t]
\vspace{-0.18in}
  \centering
  \subfloat[128 samples/minibatch]{\label{fig:bs128act}
    \includegraphics[width=0.3\linewidth]{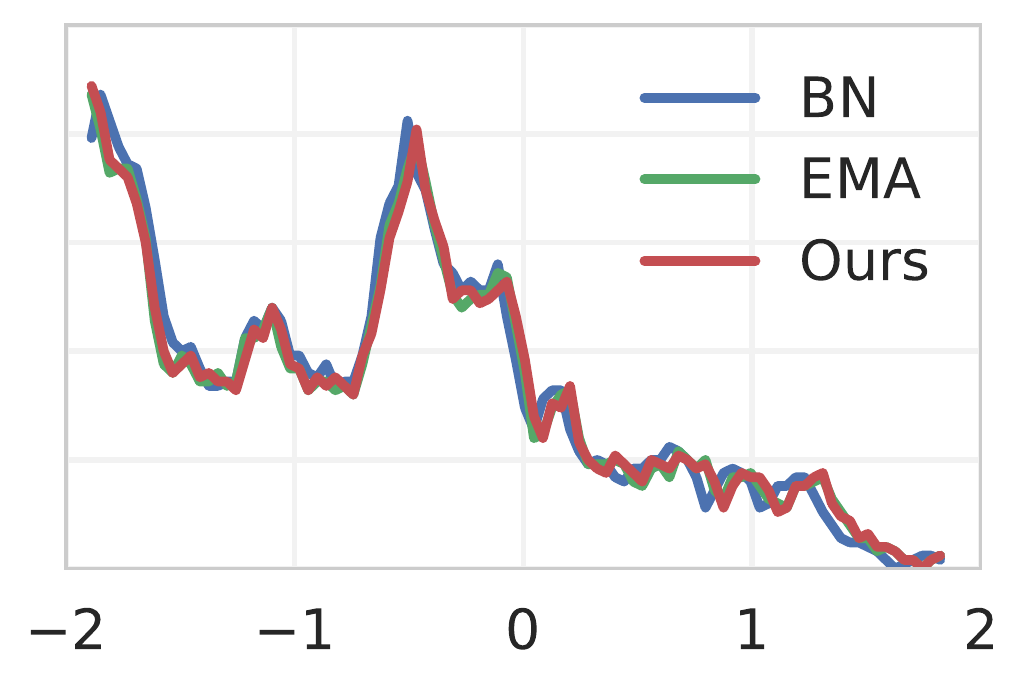}
  }
  \subfloat[2 samples/minibatch]{\label{fig:bs2act1}
    \includegraphics[width=0.3\linewidth]{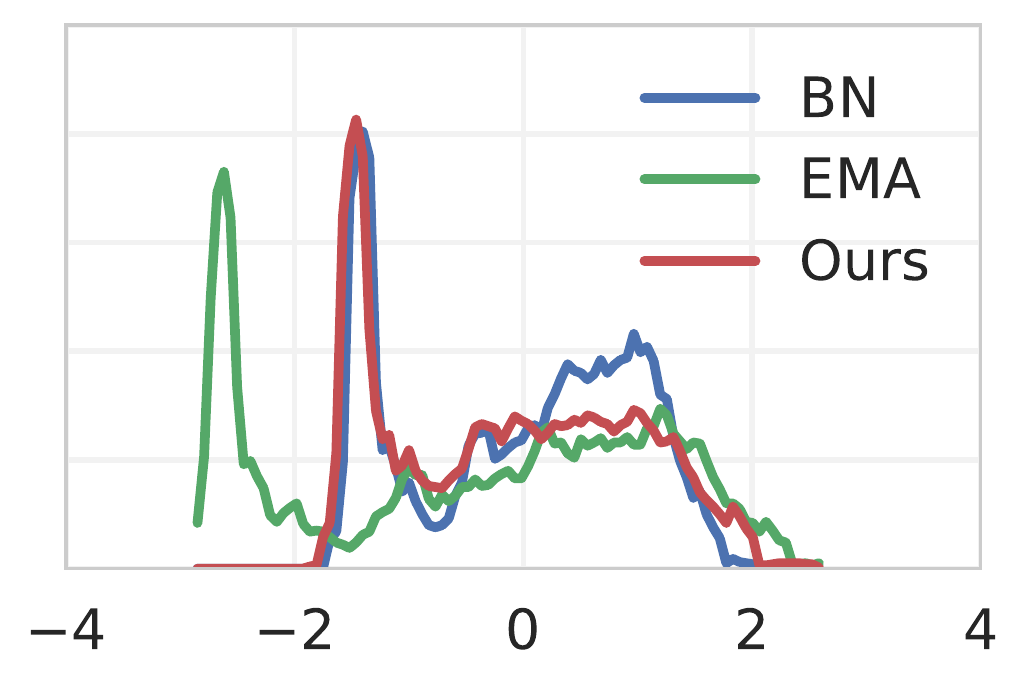}
  }
  \subfloat[2 samples/minibatch]{\label{fig:bs2act2}
    \includegraphics[width=0.3\linewidth]{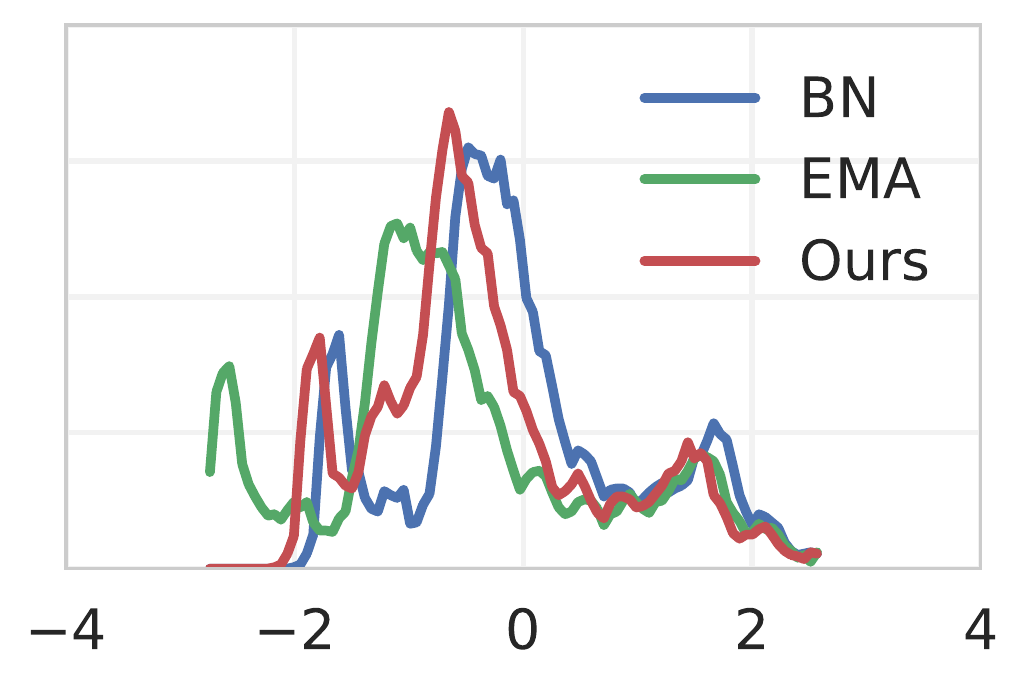}
  }
  \vspace{-0.2cm}
  \caption{
  \textbf{Discrepancy in train vs. test distributions:} In each plot we show normalized activation distributions using different normalization operations and different normalization minibatch sizes for an arbitrary channel of a ResNet-20 trained on CIFAR-100. `BN' denotes result of batch normalization during \emph{training}, `EMA' denotes normalization used during \emph{evaluation} using EMA, and `Ours' is using our EvalNorm adjustment during \emph{evaluation}. Notice that for larger minibatches (128 samples) in (a), all three normalization operation result in similar distributions. However, for small minibatches (2 samples) in (b) and (c), `Ours' using test time EN is much closer to the train time `BN' than the standard `EMA' statistics.}
  \vspace{-0.4cm}
  \label{fig:activations}
\end{figure*}

Batch Normalization (BN)~\cite{batchnorm} has significantly contributed to the wide application and success of deep learning. It has enabled training of larger and deeper models~\cite{resnets, resnetsv2, googlenet, szegedy2017inception} leading to significant advances in computer vision. 
However, a well-acknowledged drawback of BN is that it requires sufficiently large minibatches~\cite{batchrenorm,szegedy2017inception,resnets}, and results in drastically reduced model performance for smaller mini-batches.
In this paper we study this peculiar behavior of BN to gain a better understanding of the source of the problem. We identify a cause and propose \emph{EvalNorm} to address the issue. EvalNorm estimates corrected normalization statistics to use for BN during evaluation only.
EvalNorm supports online estimation of the corrected statistics while the model is being trained and does not affect the training scheme of the model. 
As a result, EvalNorm can also be used with existing pre-trained models allowing them to benefit from our method without retraining. For pre-trained models EvalNorm suggests: 1) a rule of thumb that is trivial to implement and, 2) an offline estimation method that requires a pass through data (but doesn't update the model). The ability to estimate corrected statistics online for new models helps avoid an otherwise two step process. The model is still trained as it would be without our method. Note that proposing a new normalization technique is not a goal of this paper. Instead, the goal is to gain a better understanding of the behavior of BN for small minibatches. We limit ourselves to explorations that only affect the evaluation of a model trained with BN and refer to our method as `Eval Normalization' (EvalNorm or EN).

Reliance of BN on large minibatches is prohibitive in several settings due to the hardware memory limitations. For example, applications requiring high-resolution inputs (object detection, segmentation, medical image analysis, etc.) or high-capacity (deeper and wider) networks are constrained to using smaller minibatches and thus take a performance hit.
As a result, several normalization techniques, such as Group Normalization~\cite{groupnorm} and Batch Re-normalization~\cite{batchrenorm}, have been proposed to address the problem due to smaller minibatches. However, these approaches don't shed any light on the source of the problem with BN. Instead, they propose alternatives that make changes to the model and require retraining. As a result, many already trained and deployed models that use BN, where retraining is not possible, can not benefit from these alternatives.
Our experimental evaluation of EN shows that it addresses the shortcomings of BN for small batches and provides a feasible alternative when retraining isn't an option.

\medskip
\noindent
\textbf{Training and evaluation discrepancy in BN:} During training, BN normalizes each channel for an example using the mean and variance of that channel aggregated across the full minibatch. This aggregation across the minibatch introduces a dependency on other minibatch samples. However, during evaluation each example is evaluated by itself and thus an approximation of the minibatch statistics is required.
Typically, an exponential moving average (EMA) of minibatch moment statistics is maintained during training and used as a substitute during evaluation. Normalization using EMA statistics is assumed to provide an accurate approximation to normalization observed during training.
While reasonable for larger minibatches, we demonstrate that this assumption is erroneous for small minibatches (\Cref{sec:stochsource}). This is because the mean and variance used to normalize a sample in a minibatch during training depend on that sample itself. For small minibatches this dependency is significant but ignored by the default method of using EMA. We qualitatively illustrate the discrepancy in train and test normalization in \Cref{fig:activations}, where we plot the distribution of normalized features at train (`BN') and test (`EMA') time. Each plot shows smoothed histograms of activations for an arbitrary channel of an arbitrary layer (in a ResNet-20 model) using different methods. For larger minibatches (\Cref{fig:bs128act}), we observe that the distribution of normalized activations during training and testing match well. However, for small minibatches (\Cref{fig:bs2act1,fig:bs2act2}), normalized feature distributions are quite different. Lastly, note that our method (`Ours') reduces this discrepancy and brings the distribution of normalized activations during testing closer to that during training (`BN').

To summarize our contributions, we: 1) quantify the dependence of the normalization statistics used by BN on a particular minibatch sample leading to an insight into the source of poor small minibatch performance, 2) propose two methods for estimating a correction without retraining existing models that use BN, and 3) propose a method to estimate corrections online during training of new models without affecting the training. We have included an extensive experimental section that analyzes and validates our insight and demonstrates that our insight leads to an improved evaluation performance of BN on a variety of common benchmarks.

\section{Related work}

Normalization of data for training is regularly used in machine learning. For example, it's a common practice to normalize features (scale or whiten) before learning classifiers like SVM. Similarly, for deep networks, many techniques have been proposed to normalize both inputs and intermediate representations, which make the training better and faster~\cite{lecun1998efficient,glorot2010understanding}. Batch Normalization or BatchNorm (BN) is one such technique which aims to stabilize latent feature distributions in a deep network. BN normalizes features using the statistics computed from a minibatch; and has been shown to ease the learning problem and enable fast convergence of very deep network architectures. However, with BN's widespread adoption, it has been observed that models using BN exhibit severe degradation in performance when trained with smaller minibatches (as discussed in \Cref{sec:intro}).

To address the stochasticity due to small minibatches and bias due to non-iid samples, \citet{batchrenorm} introduced batch renormalization (Renorm) which constrains the minibatch moments to a specific range. This limits the variation in minibatch statistics during training. \citet{batchrenorm} also introduced hyperparameters to prevent drift in the EMA statistics due to the variability of small minibatches. However, Renorm is still dependent on minibatch statistics with small minibatches, leading to worse performance. For tasks where small minibatches are standard (e.g., object detection), another approach is to engineer systems that can circumvent the issue. \citet{peng2017megdet} proposed to perform synchronized computation of BN statistics across GPUs (Cross-GPU BN) to obtain better statistics. However, since images/GPU for object detection is small (1-2 images), this approach requires ${\sim}128$ GPUs to compute reasonable BN estimates. Further, this does not address the original problem of BN with small minibatches. Moreover, the need for synchronized computation prohibits the use of asynchronous training, which is a standard and practical tool for large-scale problems.

Instead of dealing with the small minibatch problem, several normalization techniques have been proposed that do not utilize the construct of a `minibatch.' Instance Normalization~\cite{UlyanovVL16} performs normalization similar to BN but only for a single sample and was shown to be effective on image style transfer applications. Similarly, Layer Normalization~\cite{ba2016layer} utilizes the entire layer (all channels) to estimate the normalization statistics.
These approaches~\cite{ba2016layer,UlyanovVL16} have not shown benefits on image recognition tasks, which is the application we focus on. Instead of normalizing the activations, Weight Normalization~\cite{salimans2016weight} reparameterizes the weights in the neural network to accelerate convergence. 
Normalization Propagation~\cite{arpit2016normalization} uses data independent moment estimates in every layer, instead of computing them from minibatches during training. Group Normalization (GN)~\cite{groupnorm} divides the channels into groups and, within each group, computes the moments for normalization. GN alleviates the small minibatch problem to some extent, but it performs worse than BN for larger minibatches. \citet{ren2016normalizing} provide a unifying view of the different normalization approaches by characterizing them as the same transformation but along different dimensions (layers, samples, filters, etc.). 

Unlike many approaches discussed above, the proposed EN does not modify the training scheme of batch normalization. It only estimates a different set of evaluation time statistics. The parameters used in EN are independent of the deep network, and training the parameters does not impact the network's training in any way. We conjecture that EN may be complementary to some of the above-mentioned approaches and may be used in conjunction with them to normalize across batch dimension. However, we consider this beyond the focus of this study.

\section{Eval Normalization (EN)}\label{sec:approach}
We first briefly describe the relevant aspects of batch normalization and then present our method.

\subsection{Batch Normalization (BN)}
\label{sec:bn}
BN normalizes a particular channel of a sample in a minibatch using the mean and variance of that channel computed across the whole minibatch. Since the normalization of the channels is decoupled, we analyze the normalization of a single (but arbitrary) channel. Consider the set of activations $\BB = \{\vect{x_1}, \dots, \vect{x_B}\}$ of a particular channel in an arbitrary layer for a minibatch of size $B$. Typically, $\vect{x_i}$ is a two dimensional field of scalars. During training, BN uses the minibatch mean $\mu_{\BB}$ and variance $\sigma_{\BB}^2$ to compute the normalized activations $\vect{\hat{x}_i}$ as 
\vspace{-0.055in}
\begin{align}
\vect{\hat{x}_i} = \frac{(\vect{x_i} - \mu_{\BB})}{\sigma_{\BB}}.
\end{align}
This has two notable ramifications during training: 1) For activations $\vect{x_i}$ the normalization statistics $\mu_{\BB}$ and $\sigma_{\BB}^2$ are a function of the statistics of activations $\vect{x_i}$ themselves, 2) the normalized activations $\vect{\hat{x}_i}$ for a particular sample are stochastic as they depend on the statistics of the other samples in a stochastic minibatch. 

During evaluation randomness of $\vect{\hat{x}_i}$ due to stochastic dependency on other minibatch elements poses a difficulty for BN. To keep the evaluation deterministic and remove the dependency on other test samples BN substitutes $\vect{\hat{x}_i}$ by an estimate of its expected value $\EXP[\vect{\hat{x}_i}]$. This is done by treating $\mu_{\BB}$ and $\sigma_{\BB}^2$ encountered during training as random variables and substituting them by an estimate of their expected values as $\mu_{\EE} \approx \EXP[\mu_{\BB}]$ and $\sigma_{\EE}^2 \approx \EXP[\sigma_{\BB}^2]$ to construct a first order approximation of $\EXP[\vect{\hat{x}_i}]$ as
\begin{align}
\EXP[\vect{\hat{x}_i}] \approx \frac{(\vect{x_i} - \EXP[\mu_{\BB}])}{\sqrt{\EXP[\sigma_{\BB}^2]}} \approx \frac{(\vect{x_i} - \mu_{\EE})}{\sqrt{\sigma_{\EE}^2}}.
\label{eq:bnapprox}
\end{align}
The estimates $\mu_{\EE}$ and $\sigma_{\EE}^2$ are typically maintained as exponential moving averages (EMA) during training.

Note that BN also employs a learned affine transform after normalization. We omit this affine transform in the text as it is not relevant to the discussion of normalization.

\begin{figure*}[t]
  \centering
  \subfloat[\cifarten]{\label{fig:cifar10_baseline}
  \includegraphics[width=0.43\linewidth]{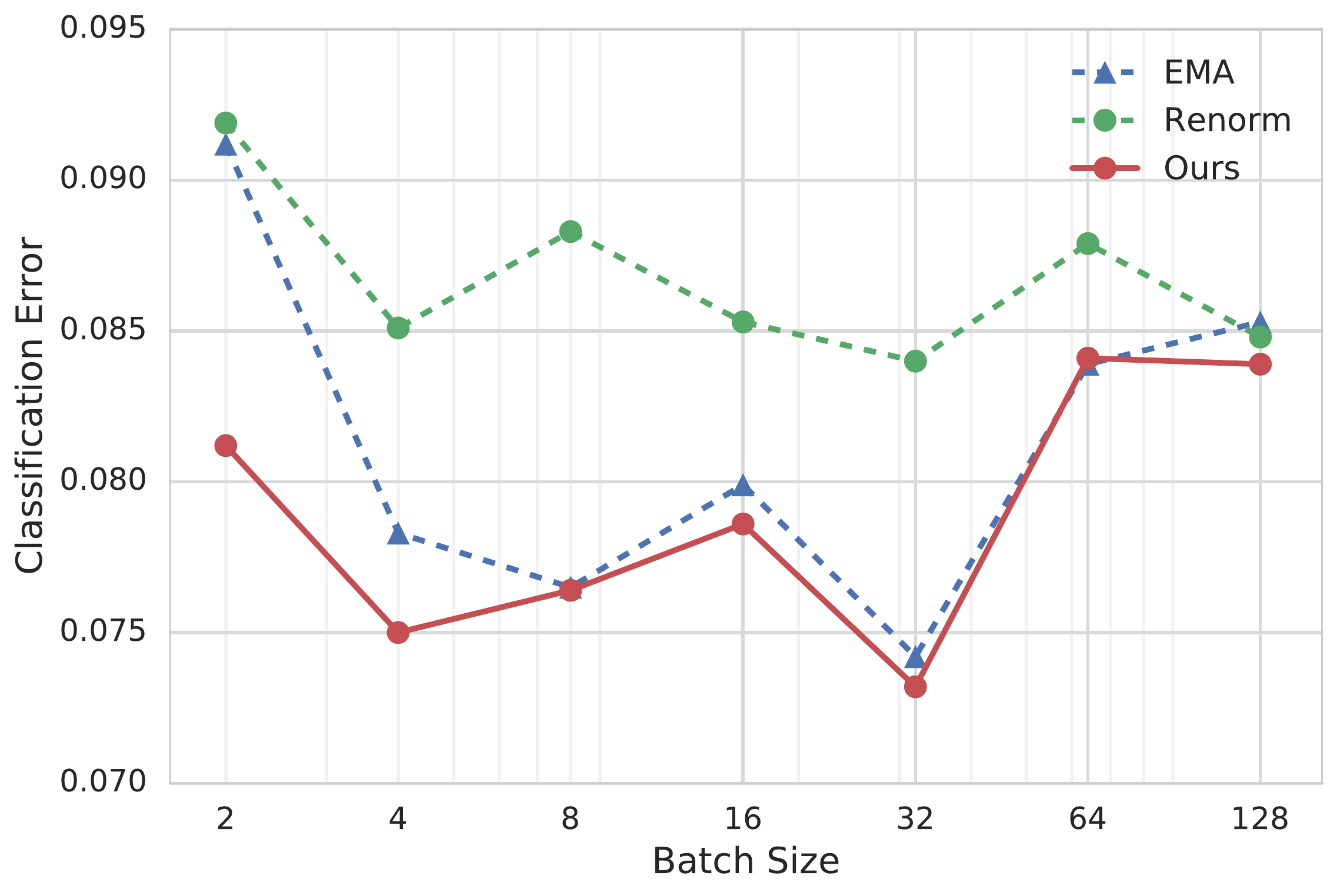}
  }
  \;
  \subfloat[\cifarhun]{\label{fig:cifar100_baseline}
  \includegraphics[width=0.43\linewidth]{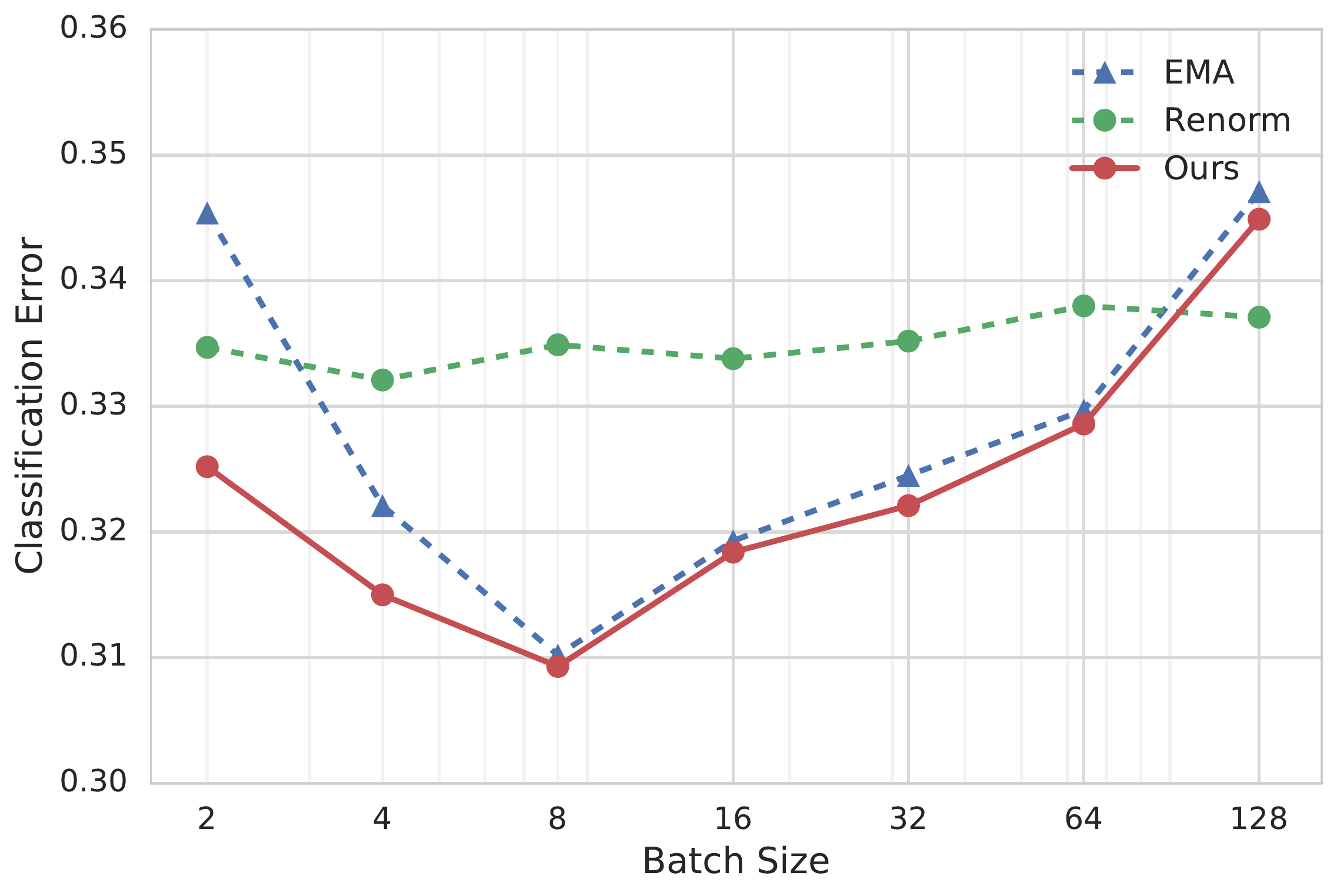}
  }
  \vspace{-0.2cm}
  \caption{\textbf{Classification error on \cifarten and CIFAR-100}. EN consistently outperforms both EMA and Renorm for all minibatch sizes. Notice that the error does not decrease monotonically with increasing minibatch sizes. One intuition behind this is that at the inflection point, the stochasticity of the estimates act as a good regularizer (similar trends were observed by~\citet{smallBatchTraining}).}
   \vspace{-0.4cm}
  \label{fig:cifar_results}
\end{figure*}

\subsection{Source of stochasticity in $\vect{\hat{x}_i}$}
\label{sec:stochsource}
As noted earlier, $\mu_{\BB}$ and $\sigma_{\BB}^2$ are a function of $\vect{x_i}$. However, \cref{eq:bnapprox} ignores this dependency entirely. Let us first make this dependency explicit. 
Let $\mu_i$ and $\sigma_i^2$ be the mean and variance respectively of $\vect{x_i}$. Denote $\CC = \BB - \vect{x_i}$ as the set of $B-1$ mini-batch elements excluding $\vect{x_i}$ with combined mean $\mu_{\CC}$ and variance $\sigma_{\CC}^2$. The full minibatch mean $\mu_{\BB}$ and variance $\sigma_{\BB}^2$ can be expressed in terms of the above with the combined population mean and variance formulae. Let $\alpha=1/B$, then
\begin{align}
\mu_{\BB}      &= \alpha \mu_i + (1-\alpha)\mu_{\CC} \label{eq:smallbatchfull1} \\
\sigma_{\BB}^2 &= \alpha\sigma_i^2 + (1-\alpha)\sigma_{\CC}^2 +\alpha(1 - \alpha)(\mu_i - \mu_{\CC})^2. \label{eq:smallbatchfull2}
\end{align}
First note that $0 < \alpha \le 1$, since $B\ge 1$. Therefore, $\alpha$ can be viewed as interpolating between the contributions from the sample being normalized and the other minibatch samples. For normalized activations $\vect{\hat{x}_i}$ it is evident that the true source of stochasticity are $\mu_{\CC}$ and $\sigma_{\CC}^2$ due to randomized minibatch. Therefore, we assert that $\EXP[\vect{\hat{x}_i}]$ in \cref{eq:bnapprox} for the minibatch sample $\vect{x_i}$ should be approximated using \cref{eq:smallbatchfull1,eq:smallbatchfull2} as opposed to approximating $\mu_{\BB}$ and $\sigma_{\BB}^2$ directly.

Note that for large minibatches $\alpha \approx 0$, resulting in a nominal contribution of $\mu_i$ and $\sigma_i^2$ to the normalizing statistics (i.e., $\EXP[\mu_{\BB}] \approx \EXP[\mu_{\CC}]$ and $\EXP[\sigma_{\BB}^2] \approx \EXP[\sigma_{\CC}^2]$). Therefore, $\EXP[\mu_{\BB}] \approx \EXP[\mu_{\CC}] \approx \mu_{\EE}$ and $\EXP[\sigma_{\BB}^2] \approx \EXP[\sigma_{\CC}^2] \approx \sigma_{\EE}^2$ (that is, using the EMA statistics for normalization) are reasonable approximations. 
However, for small minibatches the contributions of $\mu_i$ and $\sigma_i^2$ in \cref{eq:smallbatchfull1,,eq:smallbatchfull2} can not be ignored and these approximations are not accurate. We argue that this inaccuracy is the primary reason for observed distribution mis-matches between BN and EMA in \Cref{fig:activations}, and poor performance of BN for small minibatches.

\subsection{Approximating $\EXP[\vect{\hat{x}_i}]$ for small minibatches}

\noindent
\textbf{Simple approximation:}
At first glance a simple solution would be to directly use $\alpha=1/B$ and approximate $\EXP[\mu_{\BB}]$ and $\EXP[\sigma_{\BB}^2]$ using \cref{eq:smallbatchfull1,eq:smallbatchfull2}. More specifically,
$\EXP[\mu_{\BB}] \approx \alpha \mu_i + (1-\alpha)\mu_{\EE}$ and 
$\EXP[\sigma_{\BB}^2] \approx \alpha\sigma_i^2 + (1-\alpha)\sigma_{\EE}^2 +\alpha(1 - \alpha)(\mu_i - \mu_{\EE})^2$. These can then be substituted in \cref{eq:bnapprox} for normalization. Note that we have made the additional approximations of $\EXP[\mu_{\CC}] \approx \mu_{\EE}, \EXP[\sigma_{\CC}^2] \approx \sigma_{\EE}^2$.
However, for small minibatches $\mu_{\CC}$ and $\sigma_{\CC}^2$ exhibit high variance and these approximations may be inaccurate. Unfortunately, higher order approximations to account for the variance either require tracking higher order moments, whose estimates would themselves be unreliable, or making strong assumptions about the distributions of $\vect{x_i}$. Nevertheless, we explore this alternative in experiments where it turns out that $\alpha=1/B$ is less than ideal. Experimenting with a few heuristic choices for the value of $\alpha$ leads to a simple \emph{rule-of-thumb} of $\alpha=1/B^2$ (\Cref{fig:alpha_manual} and \Cref{tab:imagenet_results}) that is found to work well empirically.

\smallskip
\noindent
\textbf{Estimated approximation:}
Instead of using a fixed $\alpha$, we propose to turn $\alpha$ into an estimated variable. We instantiate two separate copies of $\alpha$, namely $\hat{\alpha}$ and $\hat{\beta}$, and construct the following approximations to \cref{eq:smallbatchfull1,eq:smallbatchfull2}
\begin{align}
\EXP[\mu_{\BB}] &\approx \hat{\alpha} \mu_i + (1-\hat{\alpha})\mu_{\EE} \label{eq:smallbatchapprox1} \\
\EXP[\sigma_{\BB}^2] &\approx \hat{\beta}\sigma_i^2 + (1-\hat{\beta})\sigma_{\EE}^2 +\hat{\beta}(1- \hat{\beta})(\mu_i - \mu_{\EE})^2. \label{eq:smallbatchapprox2}
\end{align}
Our method estimates $\hat{\alpha}$ and $\hat{\beta}$ (\Cref{sec:estimating_alpha_beta}) such that normalized activations using above approximations match the normalized activations using minibatch statistics. Note that, decoupled $\hat{\alpha}$ and $\hat{\beta}$ lead to slightly better empirical results than a single value.

\subsection{Estimating $\bm{\hat{\alpha}}$ and $\bm{\hat{\beta}}$}
\label{sec:estimating_alpha_beta}
\noindent
\textbf{Offline estimation:} Given a trained model, we propose to estimate $\hat{\alpha}$ and $\hat{\beta}$
for a particular layer by minimizing an auxiliary loss $\mathcal{L}_{\text{aux}}$ as below. Let $\stopgrad{\cdot}$ represent a \texttt{stop\_gradient} operation that does not allow flow of gradients to its argument in automatic differentiation frameworks. Then $\mathcal{L}_{\text{aux}}$ is computed as
\begin{align}
\MoveEqLeft[3.5] \hat{\mu} = \; \hat{\alpha} \stopgrad{\mu_i}+ (1-\hat{\alpha})\mu_{\EE} \\
\MoveEqLeft[4] \hat{\sigma}^2 = \; \hat{\beta}\stopgrad{\sigma_i}^2 + (1-\hat{\beta})\sigma_{\EE}^2 +\hat{\beta}(1- \hat{\beta})(\stopgrad{\mu_i} - \mu_{\EE})^2  \\ 
\MoveEqLeft[4] \mathcal{L}_{\text{aux}} = \;  \norm{\Stopgrad{\frac{\vect{x_i} - \mu_{\BB}}{\sigma_{\BB}}} - \frac{\stopgrad{\vect{x_i}} - \hat{\mu}}{\hat{\sigma}} }_1
\end{align}
Minimizing this objective allows us to estimate $\hat{\alpha}, \hat{\beta}$ such that evaluation time normalization produces activations that are similar to those observed using minibatch statistics for normalization. Further, the \texttt{stop\_gradient} operation prevents the estimation from affecting the model parameters.

\medskip
\noindent
\textbf{Online estimation:}
For training a new model a na{\"i}ve approach would be to first train the model as usual with BN and then estimate $\hat{\alpha}$ and $\hat{\beta}$ in a second pass while freezing rest of the parameters using $\mathcal{L}_{\text{aux}}$ above. However, use of \texttt{stop\_gradient} as above allows us to collapse the two steps into a single one without affecting the training of the model parameters. 

\section{Experiments}

\begin{figure*}[t]
    \centering
    \small
    \includegraphics[width=\textwidth]{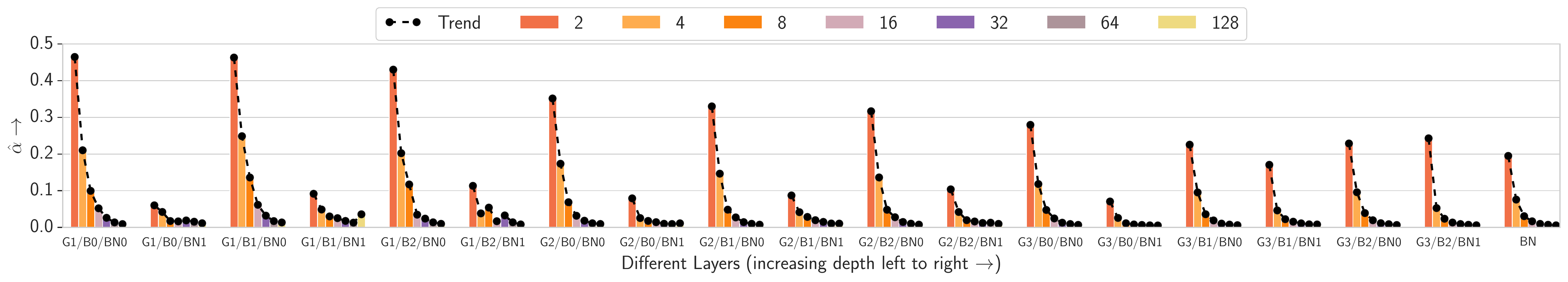}
    \includegraphics[width=\textwidth]{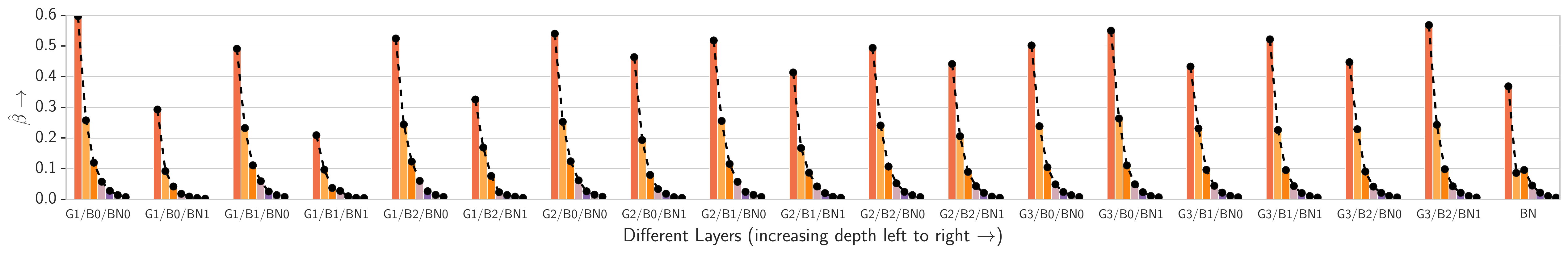}
    \vspace{-0.5cm}
    \caption{Estimated $\hat{\alpha}$ (top) and $\hat{\beta}$ (bottom) values for different batchsizes on CIFAR-100. Each set of bars corresponds to a BN operation (e.g. G1/B0  represents group 1, block 0; BN0 and BN1 represent first and second BN operation) and each color corresponds to a batch size (increasing left to right in a set). Both $\hat{\alpha}$ and $\hat{\beta}$ take larger values for smaller batchsizes indicating that individual statistics need to be included in the normalization statistics and weighted in accordance with the batchsize. }
    \vspace{-0.2cm}
    \label{fig:alphass}
\end{figure*}

We evaluate our approach on two tasks and four datasets: image classification on \cifarten, \cifarhun~\cite{krizhevsky2014cifar}, and ImageNet~\cite{imagenet}, and object detection on COCO~\cite{coco}. 
We use \cifarhun as a test-bed to perform ablation experiments and study various aspects of our method. The results on ImageNet and COCO demonstrate the utility of using EvalNorm (EN), especially for models trained with small mini-batches. Note that, unless explicitly stated, $\hat{\alpha}$ and $\hat{\beta}$ in all the experiments are estimated online for ease of experimentation.

\subsection{Image Classification on \cifarten/100}\label{sec:cifar}

\medskip
\noindent\textbf{Experimental setup.} 
\cifarten and \cifarhun contain images from 10 and 100 classes respectively.
For both \cifarten and \cifarhun, we train on the 50000 training images and evaluate on the 10000 test images. Unless specified otherwise, we use a 20 layer ResNetv2 CIFAR variant~\cite{resnetsv2} for both datasets. The base CIFAR variant contains three groups of residual blocks with widths $\{16, 32, 64\}$ respectively. Wider variants use multiples of these sizes. All networks are trained using SGD with momentum for 128k updates, with an initial learning rate of 0.1 and a cosine decay schedule~\cite{loshchilov2016sgdr} unless otherwise specified.

\begin{figure}[t]
  \centering
  \includegraphics[width=0.9\linewidth]{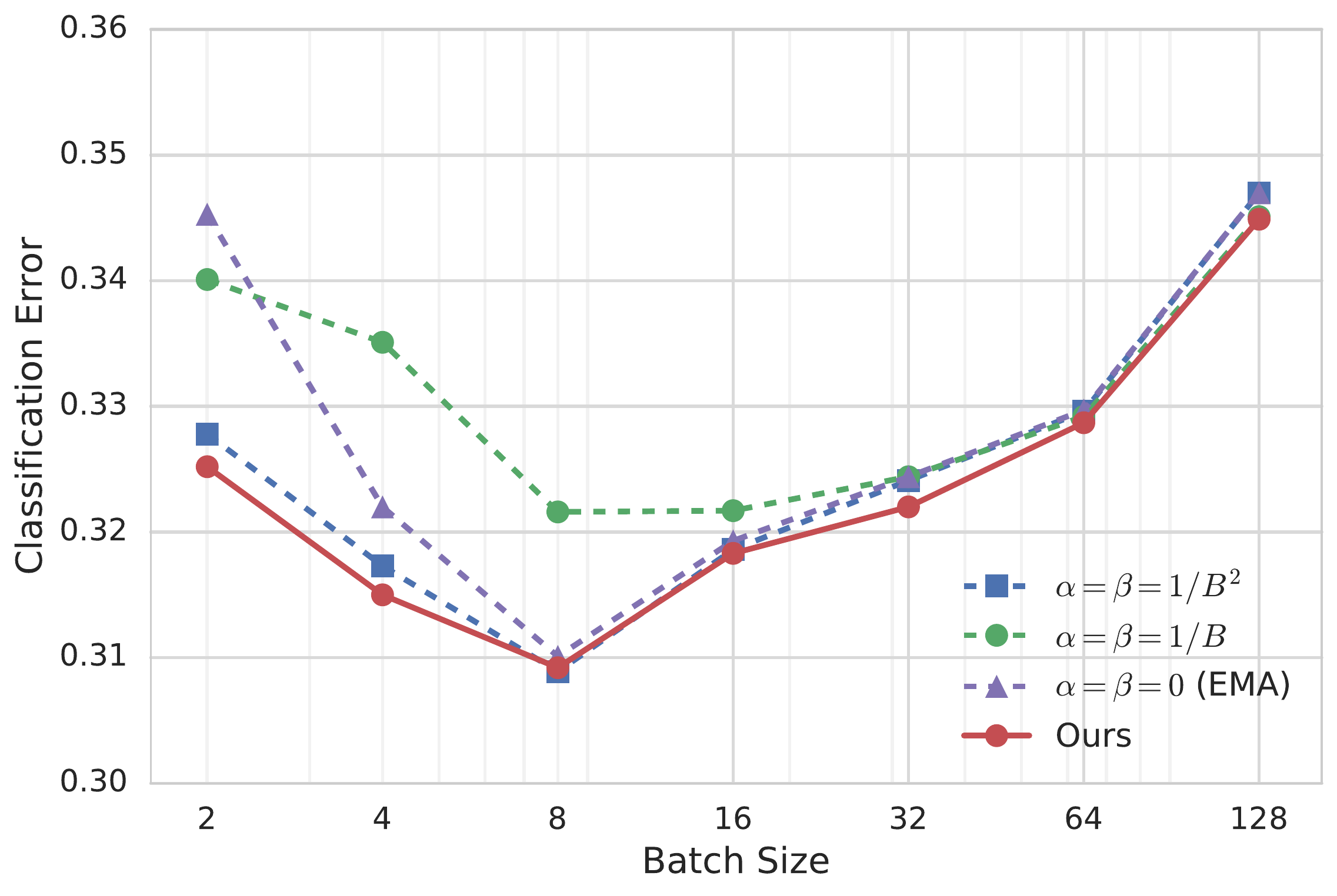}
  \vspace{-0.2cm}
  \caption{Comparison of a set of hand picked schedules for $\hat{\alpha}, \hat{\beta}$ as a function of batch size $B$. Although, the estimated values using our method perform the best, $\hat{\alpha}= \hat{\beta}=1/B^2$ appears to be a good rule of thumb. }
  \vspace{-0.1cm}
  \label{fig:alpha_manual}
\end{figure}

\begin{figure*}[t]
  \centering
  \subfloat[Depth]{\label{fig:cifar100_deeper}
    \includegraphics[width=0.44\linewidth]{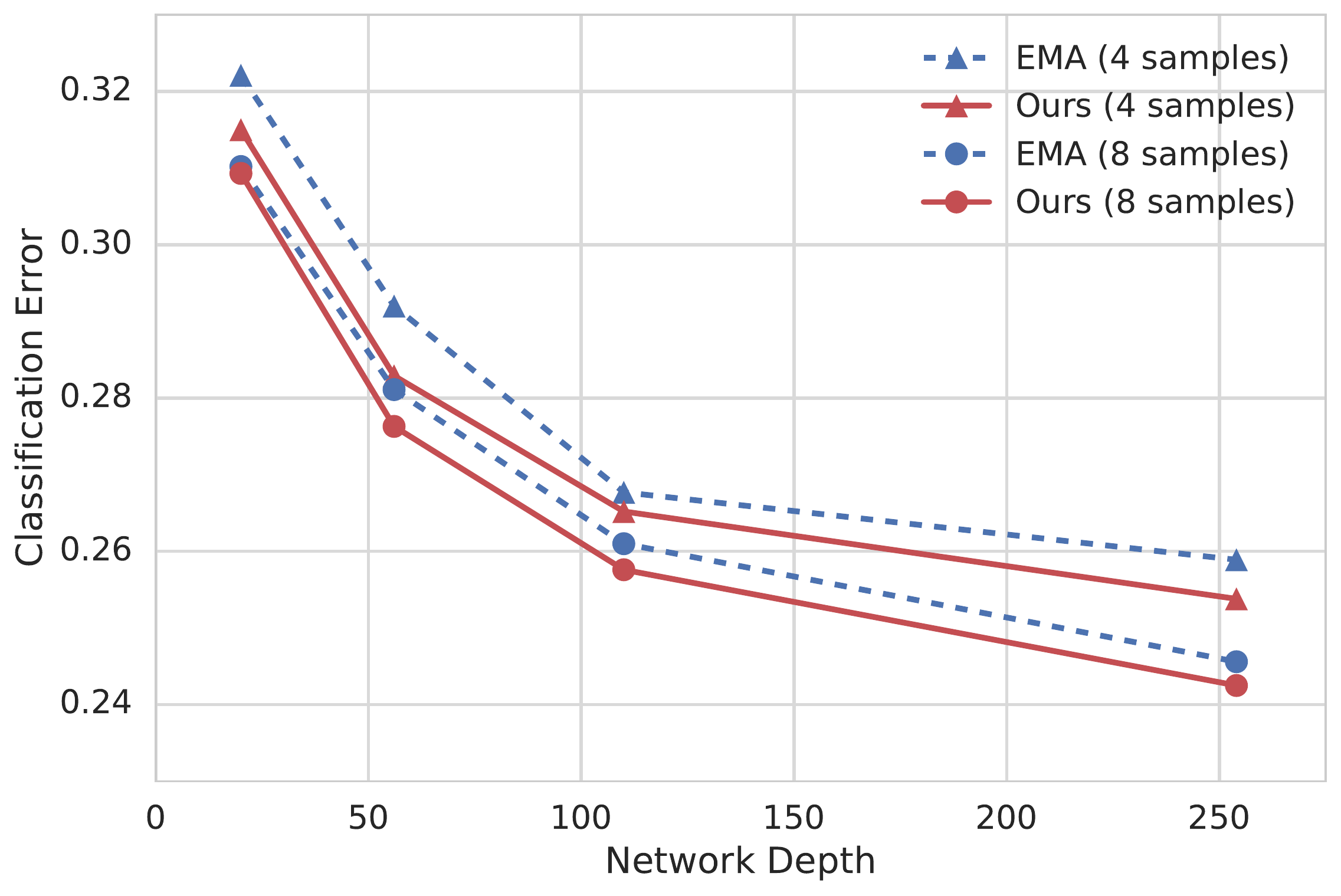}
  }
  \;
  \subfloat[Width]{\label{fig:cifar100_wider}
    \includegraphics[width=0.44\linewidth]{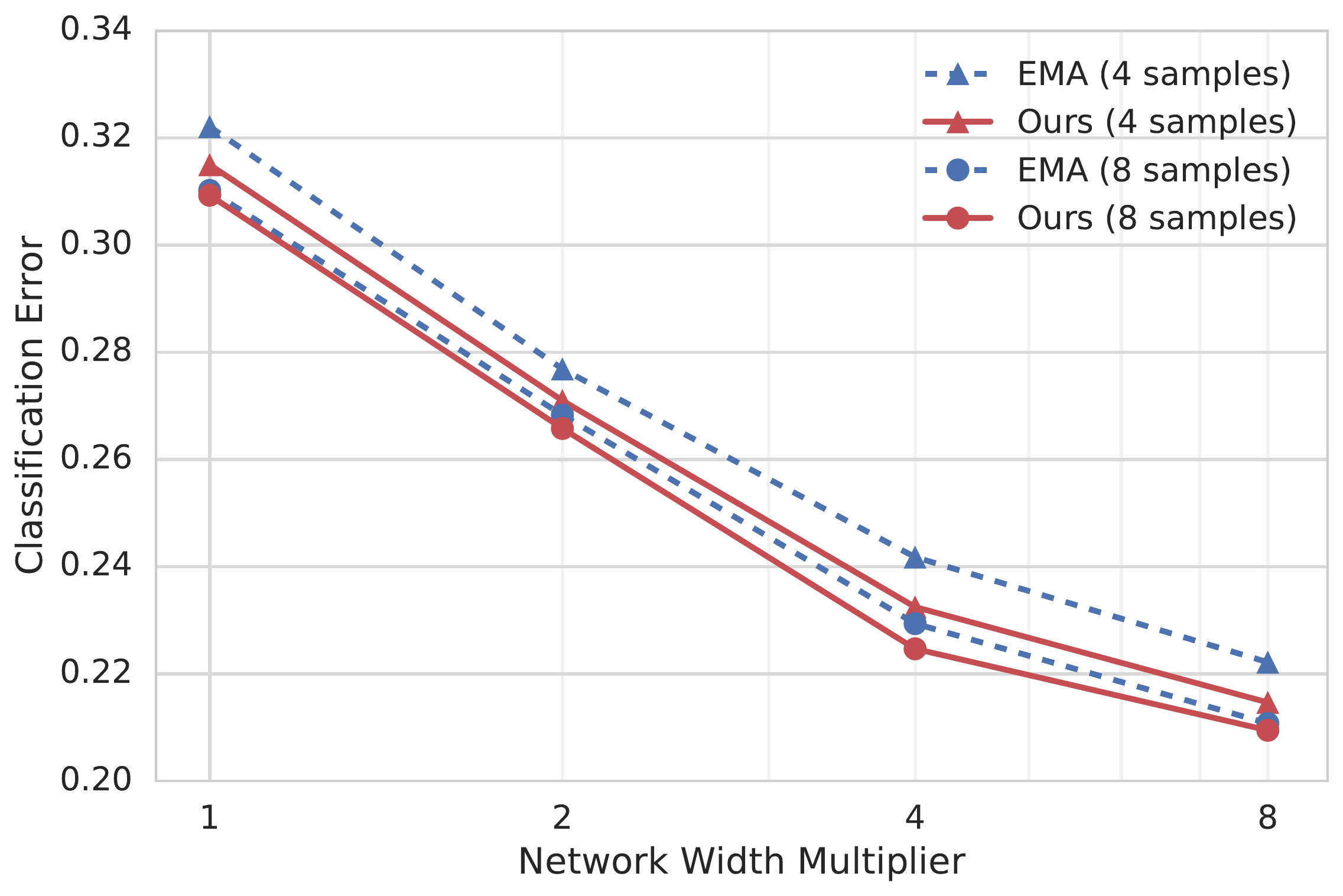}
  }
  \vspace{-0.2cm}
  \caption{Performance of EN for deeper and wider network variants on CIFAR-100. EN consistently leads to lower classification error compared to EMA statistics.}
  \vspace{-0.4cm}
  \label{fig:cifar100_variants}
\end{figure*}

\medskip
\noindent\textbf{Small minibatches.}
In this section we study the effect of minibatch size used to compute the normalization statistics. For fair comparison all models need to be trained for the same number of epochs. However, this leads to smaller minibatch models getting more gradient updates. Moreover, gradients from small minibatches have higher variance in comparison to larger minibatches requiring an adjustment in learning rate. To isolate the impact of small minibatches on normalization from these confounding factors, we use the same minibatch size (128 samples) for gradient updates and vary the number of samples used for normalization (from 2 to 128) (refer to the ``microbatch'' setup in~\cite{batchrenorm}).

\Cref{fig:cifar10_baseline,fig:cifar100_baseline} compare our method (EN) with the default EMA statistics used by standard BN~\cite{batchnorm} and batch re-normalization (Renorm)~\cite{batchrenorm}. For Renorm, we set $r_\text{max}=2.0$ and $d_\text{max}=1.0$. For larger minibatch sizes (64 and 128), we notice that all methods are comparable, but for smaller minibatch sizes (2 and 4) using EN statistics leads to a lower classification error compared to both BN and Renorm. Even though Renorm is less sensitive to varying minibatch sizes, it also lead to worse performance.

Interestingly, the performance of BN and EN does not decrease monotonically with the decreasing minibatch size used for normalization. Instead, it seems to improve before becoming worse. 
Similar trends were reported by~\cite{smallBatchTraining}. In depth study of this is beyond the scope of this paper (refer to~\cite{smallBatchTraining} for an empirical study).  One intuition is that smaller minibatches introduce stochasticity that acts as a regularizer before it starts to hurt performance.

\medskip
\noindent\textbf{Visualization of estimated $\bm{\hat{\alpha}}, \bm{\hat{\beta}}$.}
 \Cref{fig:alphass} visualizes the estimated $\hat{\alpha}, \hat{\beta}$ across different layers for various batch sizes. Larger value of $\hat{\alpha}, \hat{\beta}$ for smaller batches confirms our insight that statistics of the sample being normalized need to be included and weighted in accordance with batchsize. Both $\hat{\alpha}, \hat{\beta}$ follow the trend of taking on smaller values for larger batches indicating a decreasing influence of individual statistics for larger batches in agreement with~\cref{eq:smallbatchfull1,,eq:smallbatchfull2}.
 
\medskip
\noindent\textbf{Simple approximation for $\bm{\hat{\alpha}}, \bm{\hat{\beta}}$.}
\Cref{fig:alpha_manual} compares a set of hand picked schedules for $\hat{\alpha}, \hat{\beta}$ as a function of batch size $B$. Although, the estimated values using our method perform the best, $\hat{\alpha}= \hat{\beta}=1/B^2$ appears to be a good rule of thumb. Note that this is counter intuitive to the more natural seeming $\hat{\alpha}= \hat{\beta}=1/B$. We also validate this on ImageNet in~\Cref{tab:imagenet_results}.

\medskip
\noindent\textbf{Deeper and wider networks.} \Cref{fig:cifar100_variants} compares  EN and EMA in deeper (20, 56, 110, and 254 layers) and wider (1, 2, 4, and 8 $\times$ the original width) ResNetv2 models, with normalization minibatch size of 4 and 8 on \cifarhun. EN consistently leads to lower classification error compared to using EMA statistics indicating its general applicability.

\subsection{Image Classification on ImageNet}\label{sec:imagenet}
We report results on the ImageNet classification dataset~\cite{imagenet} with 1000 classes. We train on ${\sim}1.28$M training images and evaluate on 50k validation images. All methods in this section use a 50 layer ResNetv2~\cite{resnetsv2} network architecture. We resize all images to $229{\times}229$ and use training time data augmentation from~\cite{szegedy2017inception}.

As is standard practice~\cite{resnets,resnetsv2}, we use synchronous SGD across 8 GPUs to train all models in this section. We compute the BN and EN statistics per GPU, whereas gradients are averaged across all GPUs. Therefore, the effective minibatch size for gradient computation (SGD `batchsize') is $8\times$ the per GPU normalization minibatch size (`samples/GPU'). This is the standard synchronized multi-GPU training setup in popular libraries, such as PyTorch and Tensorflow. 

We study normalization minibatch sizes (samples/GPU) of 2, 4, 8, 16, and 32, leading to an effective SGD batchsize ($N$) of 16, 32, 64, 128, and 256 respectively. All models are trained for 60 epochs with an initial learning rate of $0.1{\times}N/256$ and a cosine decay schedule. Other implementation details (such as initialization) follows~\cite{resnetsv2}. We report results on two image classification metrics: `Prec@1' and `Recall@5'. Prec@1 measures the classification accuracy of the top-1 prediction, and Recall@5 measures the recall accuracy for top-5 predictions.

We evaluate using EN and EMA statistics for inference time normalization and report the results in \Cref{tab:imagenet_results}. We observe that using EN statistics \textit{consistently} perform better than EMA statistics across all minibatch sizes. For larger minibatches, where EMA statistics provide an accurate approximation, the difference in performance is marginal. However, for small minibatches EN statistics have a significant impact. For example, for 2 samples/GPU, Prec@1 metric improves by more than 8 points
and Recall@5 metric improves by 6.27 points.
Also note that the performance gap between small and large minibatch size is much lower with EN. For example, reducing samples/GPU from 32 to 4, Prec@1 for EMA drops from 75.98 to 71.40 (-4.58 points), whereas for EN, it only drops by 2.7 points.

Next, in \Cref{tab:imagenet_results} we also compare our method with two recent normalization methods namely Group normalization~\cite{groupnorm} and Batch renormalization~\cite{batchrenorm} that circumvent minibatch dependence. Note that these methods change the model or the training method itself. EN performs similar to or better than alternatives that normalize across batch (EMA and Renorm), indicating that it properly accounts for individual sample contributions. For large batch sizes EN outperforms other methods while for small batch sizes Group normalization tends to perform better. However, the gain from retraining using GroupNorm is reduced from 10\% to 3.75\% (for batchsize 2), and EN does not suffer the side effect of reduced performance for large batch sizes. 

\Cref{tab:imagenet_results} also reports the performance using the suggested rule-of-thumb ($\alpha=\beta=1/B^2$) as well as offline estimation. Rule-of-thumb significantly improves over EMA but under-performs EN. Nevertheless, it is an attractive and easier alternative to estimation at the cost of performance. Offline estimation performs similar to online estimation as expected.
 
\begin{table}[t]
  \caption{\textbf{ImageNet classification results} for ResNet-50~\cite{resnetsv2} using EMA and EN statistics for different normalization minibatch sizes (samples/GPU). We show the classification accuracy for top-1 prediction (Prec@1) and the recall for top-5 prediction (Recall@5). $\Delta$ represents improvement due to EN over EMA. EN consistently performs better than EMA across all minibatch sizes, and provides significant gains for small minibatches. See \cref{sec:imagenet} for details.}
  \vspace{-0.2cm}
  \label{tab:imagenet_results}
  \centering
  \renewcommand{\arraystretch}{1.0}
  \renewcommand{\tabcolsep}{1.3mm}
  \resizebox{\linewidth}{!}{
  \begin{tabular}{@{}L{0.25cm} L{3.2cm} c c c c c c c@{}}
    \toprule
    \multicolumn{2}{r}{\small samples/GPU $\rightarrow$} & 32 & 16 & 8 & 4 & 2 \\
    \midrule
    \multirow{7}{*}[-2pt]{\rotatebox{90}{Prec@1}} & Renorm & 76.04 & 75.78 & 74.21 & 73.33 & 70.87 \\
                   & Groupnorm & 75.87 & 75.73 & 75.75 & 75.61 & 74.78 \\
     \cmidrule(l{0.3em}){2-7}
                   & EMA & 75.98 & 75.26 & 73.68 & 71.40 & 64.80 \\
                   & EN [\textbf{ours}] & 76.20 & 75.89 & 74.87 & 73.49 & 70.98 \\
                   & $\Delta$ & +0.22 & +0.63 & +1.19 & +2.09 & \textbf{+6.18} \\
    \cmidrule(l{0.3em}){2-7}
                   & $\alpha=\beta=1/B^2$ [\textbf{ours}] & 76.28 &	75.77 &	74.52 &	73.17 &	70.12 \\
                   & EN-Offline [\textbf{ours}] & 76.18	& 75.85	& 74.92	& 73.48 & 71.03 \\
    \midrule
    \multirow{7}{*}[-2pt]{\rotatebox{90}{Recall@5}} & Renorm & 92.81 & 92.65 & 92.18 & 91.68 & 90.03 \\
                   & Groupnorm & 92.59 & 92.31 & 92.28 & 92.10 & 91.81 \\
     \cmidrule(l{0.3em}){2-7}
                   & EMA & 92.88 & 92.61 & 92.17 & 90.70 & 86.13 \\
                   & EN [\textbf{ours}] & {92.97} & 92.78 & 92.38 & 91.75 & 90.18 \\
                   & $\Delta$ & +0.09 & +0.17 & +0.21 & +1.05 & \textbf{+4.05} \\
    \cmidrule(l{0.3em}){2-7}
                   & $\alpha = \beta = 1/B^2$ [\textbf{ours}] & 92.96 &	92.81 &	92.17 &	91.46 &	89.51 \\
                   & EN-Offline [\textbf{ours}] & 92.93	& 92.79	& 92.30 & 91.81 & 90.25 \\
    \bottomrule
  \end{tabular}
  }
  \vspace{-0.2cm}
\end{table}

\begin{table*}[t]
  \caption{\textbf{Object detection results on COCO} using Faster R-CNN~\cite{fasterrcnn} with ResNet50~\cite{resnets}. We report EMA and EN performance for different minibatch sizes (images/batch). EN performs better across all AP metrics and all minibatch sizes, specially for 2 images/minibatch, where we see significant gains. Different blocks (a)-(d) are described in \Cref{sec:coco_exp}.  ($\Delta>1$ point are shown in bold)\vspace{-0.15cm}}
  \label{tab:coco_results}
  \centering
  \renewcommand{\arraystretch}{1.0}
  \renewcommand{\tabcolsep}{1.5mm}
  \small
  \resizebox{\linewidth}{!}{

  \begin{tabular}{@{}
      C{0.5cm}
      L{1.25cm} C{0.9cm} L{.1cm} L{1.25cm} 
      C{0.9cm} L{.01cm} L{1.6cm} 
      !{\color{gray}\vrule} ccc 
      !{\color{gray}\vrule} ccc 
      !{\color{gray}\vrule} ccc
  @{}}
    \toprule
    & 
    \multicolumn{2}{@{}c@{}}{\footnotesize Network Weights} & &
    \multicolumn{2}{@{}c@{}}{\footnotesize BN EMA Stats} & & & 
    \multicolumn{3}{@{}c|@{}}{AP} &  
    \multicolumn{3}{@{}c|@{}}{AP$^{{50}}$} & 
    \multicolumn{3}{@{}c@{}}{AP$^{{75}}$} 
    \\
    \cmidrule(r{0em}){2-3} \cmidrule(r{0em}){5-6}  \cmidrule(l{0.2em}){8-17}
    & \footnotesize{Init} & \footnotesize{Train?} & & \footnotesize{Init} & \footnotesize{Train?} & 
    \multicolumn{2}{@{}r@{}}{\small \color{gray}{images/batch} $\rightarrow$} & 8 & 4 & 2 & 8 & 4 & 2 & 8 & 4 & 2 \\
    \midrule
    \multirow{3}{*}[-2pt]{\small \color{red}{(a)}} & 
    \multirow{3}{*}[-2pt]{\footnotesize Random} & 
    \multirow{3}{*}[-2pt]{\large $\checkmark$} & & 
    \multirow{3}{*}[-2pt]{\footnotesize Random} & 
    \multirow{3}{*}[-2pt]{\large $\checkmark$} &  & 
    \small{EMA} & 
    25.2 & 25.6 & 22.0 & 40.9 & 41.2 & 36.1 & 26.7 & 27.5 & 22.4\\
    & & & & & & & 
    \small{EN [\textbf{ours}]} & 
    25.2 & 25.8 & 23.5 & 41.0 & 41.3 & 38.1 & 26.8 & 27.6 & 25.0 \\
    \cmidrule(l{0.3em}){8-17}
    & & & & & & & 
    \small{$\Delta$} & 
    0.0 & +0.1 & \textbf{+1.5} & +0.1 & +0.1 & \textbf{+2.0} & +0.1 & +0.1 & \textbf{+2.6} \\
    \midrule
    \multirow{3}{*}[-2pt]{\small \color{red}{(b)}} & 
    \multirow{3}{*}[-2pt]{\footnotesize ImageNet} & 
    \multirow{3}{*}[-2pt]{\large $\checkmark$} & & 
    \multirow{3}{*}[-2pt]{\footnotesize Random} & 
    \multirow{3}{*}[-2pt]{\large $\checkmark$} &  & 
    EMA & 
    30.4 & 29.9 & 27.6 & 48.4 & 47.2 & 43.5 & 32.8 & 31.9 & 29.6 \\
    & & & & & & &
    EN [\textbf{ours}] &
    30.7 & 30.5 & 29.6 & 48.5 & 47.9 & 46.5 & 33.0 & 32.8 & 31.3 \\
    \cmidrule(l{0.3em}){8-17}
    & & & & & & & 
    \small{$\Delta$} & 
    +0.2 & +0.6 & \textbf{+2.0} & +0.1 & +0.7 & \textbf{+3.1} & +0.2 & +0.9 & \textbf{+1.7} \\
    \midrule
    \multirow{3}{*}[-2pt]{\small \color{red}{(c)}} & 
    \multirow{3}{*}[-2pt]{\footnotesize ImageNet} & 
    \multirow{3}{*}[-2pt]{\large $\checkmark$} & & 
    \multirow{3}{*}[-2pt]{\footnotesize ImageNet} & 
    \multirow{3}{*}[-2pt]{\large $\times$} &  & 
    EMA &
    27.5 & 28.8 & 28.2 & 45.1 & 47.3 & 46.6 & 28.9 & 30.5 & 30.5 \\
    & & & & & & &
    EN [\textbf{ours}] & 
    30.2 & 30.3 & 30.5 & 48.2 & 48.6 & 48.7 & 32.2 & 32.1 & 32.7 \\
    \cmidrule(l{0.3em}){8-17}
    & & & & & & &
    \small{$\Delta$} & 
    \textbf{+2.7} & \textbf{+1.5} & \textbf{+2.3} & \textbf{+3.1} & \textbf{+1.3} & \textbf{+2.1} & \textbf{+3.3} & \textbf{+1.6} & \textbf{+2.2} \\
    \midrule
    \multirow{3}{*}[-2pt]{\small \color{red}{(d)}} & 
    \multirow{3}{*}[-2pt]{\footnotesize ImageNet} & 
    \multirow{3}{*}[-2pt]{\large $\checkmark$} & & 
    \multirow{3}{*}[-2pt]{\footnotesize ImageNet} & 
    \multirow{3}{*}[-2pt]{\large $\checkmark$} &  &  
    EMA &
    31.7 & 30.1 & 24.8 & 50.6 & 47.6 & 40.0 & 33.9 & 32.2 & 26.1 \\
    & & & & & & & EN [\textbf{ours}] & 
    31.9 & 31.6 & 30.2 & 50.7 & 49.6 & 46.9 & 34.1 & 34.0 & 31.0 \\
    \cmidrule(l{0.3em}){8-17}
    & & & & & & & 
    \small{$\Delta$} & 
    +0.2 & \textbf{+1.5} & \textbf{+5.4} & +0.2 & \textbf{+2.1} & \textbf{+7.0} & +0.2 & \textbf{+1.8} & \textbf{+4.9} \\
    \bottomrule
  \end{tabular}
}
\vspace{-0.3cm}
\end{table*}

\subsection{Object Detection on COCO}\label{sec:coco_exp}
Finally, we evaluate our method on the task of object detection, and demonstrate consistent improvements when using EN statistics as opposed to EMA. Object detection frameworks are typically trained with high resolution inputs, and hence use small SGD minibatch sizes. As a result object detection is a good benchmark to evaluate the differences between EN and EMA.

\medskip
\noindent\textbf{Experimental setup.} 
All experiments in this section use the COCO dataset~\cite{coco} with 80 object classes. We train using  the trainval35k set~\cite{fastrcnn} with ${\sim}$75k images and evaluate on a held out minival set~\cite{fastrcnn} with 5k images. We report the standard COCO evaluation metrics for mean average prevision with varying IoU thresholds (AP, AP$^{{50}}$, AP$^{{75}}$) (see~\cite{coco} for details).

We use the Faster R-CNN (FRCN)~\cite{fasterrcnn,huang2016speed} object detection framework, built with a 50 layer ResNetv1~\cite{resnets} (ResNet-50). The FRCN system has three components: 1) backbone feature extractor (base network), which operates on high resolution images (we resize images to 600${\times}$600 for all experiments), 2) region proposal network (RPN), which proposes regions of interest (ROIs), and 3) region classification network (RCN), which classifies regions proposed by RPN using cropped features from the base network. All but the last residual blocks from ResNet-50 are used as the base network and the last residual block is used in the RCN network. Refer to~\cite{fasterrcnn,huang2016speed,ohem} for architecture details.

We study the SGD minibatch sizes (images/GPU or $N$) of 2, 4, and 8. As is standard practice~\cite{huang2016speed}, we use minibatch size of 300 and 64$N$ ROIs for the RPN and RCN respectively. Note that this results in different normalization minibatches for different BN layers in the network ($N$ for base network and $64N$ for RCN). All models are trained for 64 epochs (in terms of images and not ROIs) with an initial learning rate of $0.015{\times}N/8$ and a cosine decay schedule. All methods use asynchronous SGD with 11 parallel GPU workers. We use the publicly released code from~\cite{huang2016speed}.

We investigate two different training paradigms: training from a random initialization and transfer learning (or fine-tuning). The random initialization setup is similar to the experiments reported on image classification in \Cref{sec:imagenet,sec:cifar}. The transfer learning setup is more common in practice because it generally leads to higher performance. Next, we discuss the trade-offs and training flexibility in each paradigm and report results in \Cref{tab:coco_results} (blocks (a)-(d)).

\medskip
\noindent\textbf{Random initialization.}
We study the impact of EN by training a ResNet-50 Faster R-CNN model with all parameters initialized randomly. The results for using both EMA and EN statistics are reported \Cref{tab:coco_results}(a). We observe that when using 4 and 8 images/minibatch, both methods are on-par; but when training with 2 images/minibatch, we see consistent and significant gains of more than $1.5$ points on all AP metrics. Note that in this setup, the SGD and normalization minibatch sizes for the base network are same; unlike the ImageNet setup, where we average gradients across 8 GPUs resulting in SGD minibatch size being $8{\times}$ the normalization minibatch size.

\medskip
\noindent\textbf{Transfer learning.} 
The standard paradigm of training object detection models (like Faster R-CNN) is to use transfer learning or fine-tuning~\cite{rcnn,fastrcnn,fasterrcnn,huang2016speed}. In this setup, parameters from a model trained on ImageNet classification are transferred to the detection model, which is then trained on object detection. We use the model checkpoint from~\cite{resnets} as is standard practice. The transfer learning paradigm allows us to study the impact of EN statistics compared to EMA in a variety of settings.

First, we investigate which method is able to better estimate BN statistics in the absence of a prior. For this, we use the ImageNet trained model to initialize the network parameters (except BN parameters), but we use initial values of BN parameters from~\cite{batchnorm}; and both sets of parameters are trained simultaneously. We report the results for EN and EMA in \Cref{tab:coco_results}(b). We notice that EN is consistently better than EMA across all minibatch sizes and AP metrics. In fact, as opposed to random initialization, we see gains for both 2 and 4 images/minibatch, with the improvements for the smaller minibatch being much higher.

Next, we study the standard setup~\cite{fasterrcnn,huang2016speed} of initializing both, the network and the BN EMA parameters, using the ImageNet model. Since standard BN performs poorly with smaller minibatches~\cite{resnets,batchrenorm,groupnorm}, the BN parameters are not updated during training in this setup. The results for this setup are reported in \Cref{tab:coco_results}(c). Notice that EN performs significantly better than EMA across all minibatch sizes and AP metrics.
EN allows adjustment of statistics by using the learned features (as opposed to `stale' features from initialization).
Since this is the standard training paradigm used by almost all detection systems, these consistent and significant improvements are doubly important. 

Finally, to demonstrate the effectiveness of EN in adjusting statistics, we initialize both network and BN parameters from ImageNet, and fine-tune \textbf{both} for object detection. In practice, this setup performs poorly because BN is unable to train for small minibatches, and is generally not used (for example,~\cite{groupnorm} ignore this variant because of significant drop in performance). 
We also notice this in \Cref{tab:coco_results}(c) and (d), where the EMA performance for 2 images/minibatch drops by 3.4 AP, 6.6 AP$^{50}$, and 4.4 AP$^{75}$. Because of these drastic drops in performance, methods do not fine-tune BN parameters. Compare this to using the proposed EN statistics, which improves over EMA by \textbf{5.4} AP, \textbf{7.0} AP$^{50}$, and \textbf{4.9} AP$^{75}$. Therefore, using EN allows us to train BN parameters for smaller minibatches yielding significant improvements.

In this section, we showed that for the object detection task, which is generally trained with small minibatches, using EN statistics performs markedly better across all training setups. In \Cref{tab:coco_results}, notice that we get a healthy performance improvement for 2 images/minibatch across training setups (ranging from \textbf{+1.5} points to \textbf{+7} points). In the standard paradigm (\Cref{tab:coco_results}(c)), we improve for all minibatch sizes. Moreover, when training BN parameters for small minibatches, (\Cref{tab:coco_results}(c) and (d)), we improve both 2 and 4 images/minibatch.

\section{Conclusion}

Our goal in this work was to gain a better understanding of the problem with BN for small minibatches. We found that for models trained with small minibatches, normalization using EMA statistics during evaluation provides inaccurate approximation for normalization using minibatch statistics during training. This leads to a discrepancy between training and evaluation and is the main reason of performance degradation of BN for small batch sizes. We proposed EvalNorm, which provides a corrected normalization term for use at evaluation. EN is fully compatible with the existing pre-trained models using BN and yields large gains for models trained with smaller batches.

{\small
\medskip
\noindent\textbf{Acknowledgement.} We would like to thank Chen Sun, Sergey Ioffe, and Rahul Sukthankar for helpful discussions and comments; and Ishan Misra and Larry Davis for feedback on the draft.

\setlength{\bibsep}{0pt}
\bibliographystyle{plainnat}
\bibliography{paper}

\begin{thebibliography}{25}
\providecommand{\natexlab}[1]{#1}
\providecommand{\url}[1]{\texttt{#1}}
\expandafter\ifx\csname urlstyle\endcsname\relax
  \providecommand{\doi}[1]{doi: #1}\else
  \providecommand{\doi}{doi: \begingroup \urlstyle{rm}\Url}\fi

\bibitem[Arpit et~al.(2016)Arpit, Zhou, Kota, and
  Govindaraju]{arpit2016normalization}
Devansh Arpit, Yingbo Zhou, Bhargava~U Kota, and Venu Govindaraju.
\newblock Normalization propagation: A parametric technique for removing
  internal covariate shift in deep networks.
\newblock \emph{arXiv preprint arXiv:1603.01431}, 2016.

\bibitem[Ba et~al.(2016)Ba, Kiros, and Hinton]{ba2016layer}
Jimmy~Lei Ba, Jamie~Ryan Kiros, and Geoffrey~E Hinton.
\newblock Layer normalization.
\newblock \emph{arXiv preprint arXiv:1607.06450}, 2016.

\bibitem[Deng et~al.(2009)Deng, Dong, Socher, Li, Li, and Fei-Fei]{imagenet}
Jia Deng, Wei Dong, Richard Socher, Li-Jia Li, Kai Li, and Li~Fei-Fei.
\newblock Imagenet: A large-scale hierarchical image database.
\newblock In \emph{Computer Vision and Pattern Recognition}, pages 248--255.
  Ieee, 2009.

\bibitem[Girshick(2015)]{fastrcnn}
Ross Girshick.
\newblock {Fast R-CNN}.
\newblock \emph{Computer Vision and Pattern Recognition}, 2015.

\bibitem[Girshick et~al.(2014)Girshick, Donahue, Darrell, and Malik]{rcnn}
Ross Girshick, Jeff Donahue, Trevor Darrell, and Jagannath Malik.
\newblock Rich feature hierarchies for accurate object detection a nd semantic
  segmentation.
\newblock In \emph{Computer Vision and Pattern Recognition}, pages 580--587.
  IEEE, 2014.

\bibitem[Glorot and Bengio(2010)]{glorot2010understanding}
Xavier Glorot and Yoshua Bengio.
\newblock Understanding the difficulty of training deep feedforward neural
  networks.
\newblock In \emph{Proceedings of the thirteenth international conference on
  artificial intelligence and statistics}, pages 249--256, 2010.

\bibitem[He et~al.(2016{\natexlab{a}})He, Zhang, Ren, and Sun]{resnets}
Kaiming He, Xiangyu Zhang, Shaoqing Ren, and Jian Sun.
\newblock Deep residual learning for image recognition.
\newblock In \emph{Computer vision and pattern recognition}, pages 770--778,
  2016{\natexlab{a}}.

\bibitem[He et~al.(2016{\natexlab{b}})He, Zhang, Ren, and Sun]{resnetsv2}
Kaiming He, Xiangyu Zhang, Shaoqing Ren, and Jian Sun.
\newblock Identity mappings in deep residual networks.
\newblock In \emph{European conference on computer vision}, pages 630--645.
  Springer, 2016{\natexlab{b}}.

\bibitem[Huang et~al.(2017)Huang, Rathod, Sun, Zhu, Korattikara, Fathi,
  Fischer, Wojna, Song, Guadarrama, and Murphy]{huang2016speed}
Jonathan Huang, Vivek Rathod, Chen Sun, Menglong Zhu, Anoop Korattikara,
  Alireza Fathi, Ian Fischer, Zbigniew Wojna, Yang Song, Sergio Guadarrama, and
  Kevin Murphy.
\newblock Speed/accuracy trade-offs for modern convolutional object detectors.
\newblock In \emph{Computer Vision and Pattern Recognition}, 2017.

\bibitem[Ioffe(2017)]{batchrenorm}
Sergey Ioffe.
\newblock Batch renormalization: Towards reducing minibatch dependence in
  batch-normalized models.
\newblock In \emph{Advances in Neural Information Processing Systems}, pages
  1942--1950, 2017.

\bibitem[Ioffe and Szegedy(2015)]{batchnorm}
Sergey Ioffe and Christian Szegedy.
\newblock Batch normalization: Accelerating deep network training by reducing
  internal covariate shift.
\newblock \emph{CoRR}, abs/1502.03167, 2015.

\bibitem[Krizhevsky et~al.(2014)Krizhevsky, Nair, and
  Hinton]{krizhevsky2014cifar}
Alex Krizhevsky, Vinod Nair, and Geoffrey Hinton.
\newblock The cifar-10/100 dataset.
\newblock \emph{online: http://www. cs. toronto. edu/kriz/cifar. html}, 2014.

\bibitem[LeCun et~al.(1998)LeCun, Bottou, Orr, and
  M{\"u}ller]{lecun1998efficient}
Yann LeCun, L{\'e}on Bottou, Genevieve~B Orr, and Klaus-Robert M{\"u}ller.
\newblock Efficient backprop.
\newblock In \emph{Neural networks: Tricks of the trade}, pages 9--50.
  Springer, 1998.

\bibitem[Lin et~al.(2014)Lin, Maire, Belongie, Hays, Perona, Ramanan,
  Doll{\'a}r, and Zitnick]{coco}
Tsung-Yi Lin, Michael Maire, Serge Belongie, James Hays, Pietro Perona, Deva
  Ramanan, Piotr Doll{\'a}r, and C~Lawrence Zitnick.
\newblock Microsoft {COCO}: Common objects in context.
\newblock In \emph{European conference on computer vision}, pages 740--755.
  Springer, 2014.

\bibitem[Loshchilov and Hutter(2017)]{loshchilov2016sgdr}
Ilya Loshchilov and Frank Hutter.
\newblock {SGDR}: Stochastic gradient descent with warm restarts.
\newblock In \emph{{ICLR}}, 2017.

\bibitem[Masters and Luschi(2018)]{smallBatchTraining}
Dominic Masters and Carlo Luschi.
\newblock Revisiting small batch training for deep neural networks.
\newblock \emph{arXiv preprint arXiv:1804.07612}, 2018.

\bibitem[Peng et~al.(2018)Peng, Xiao, Li, Jiang, Zhang, Jia, Yu, and
  Sun]{peng2017megdet}
Chao Peng, Tete Xiao, Zeming Li, Yuning Jiang, Xiangyu Zhang, Kai Jia, Gang Yu,
  and Jian Sun.
\newblock Megdet: A large mini-batch object detector.
\newblock In \emph{Computer Vision and Pattern Recognition}, pages 6181--6189,
  2018.

\bibitem[Ren et~al.(2016)Ren, Liao, Urtasun, Sinz, and
  Zemel]{ren2016normalizing}
Mengye Ren, Renjie Liao, Raquel Urtasun, Fabian~H Sinz, and Richard~S Zemel.
\newblock Normalizing the normalizers: Comparing and extending network
  normalization schemes.
\newblock \emph{arXiv preprint arXiv:1611.04520}, 2016.

\bibitem[Ren et~al.(2015)Ren, He, Girshick, and Sun]{fasterrcnn}
Shaoqing Ren, Kaiming He, Ross Girshick, and Jian Sun.
\newblock Faster r-cnn: Towards real-time object detection with region proposal
  networks.
\newblock In \emph{Advances in neural information processing systems}, 2015.

\bibitem[Salimans and Kingma(2016)]{salimans2016weight}
Tim Salimans and Diederik~P Kingma.
\newblock Weight normalization: A simple reparameterization to accelerate
  training of deep neural networks.
\newblock In \emph{Advances in Neural Information Processing Systems}, pages
  901--909, 2016.

\bibitem[Shrivastava et~al.(2016)Shrivastava, Gupta, and Girshick]{ohem}
Abhinav Shrivastava, Abhinav Gupta, and Ross Girshick.
\newblock Training region-based object detectors with online hard example
  mining.
\newblock In \emph{Computer Vision and Pattern Recognition}, pages 761--769,
  2016.

\bibitem[Szegedy et~al.(2015)Szegedy, Liu, Jia, Sermanet, Reed, Anguelov,
  Erhan, Vanhoucke, and Rabinovich]{googlenet}
Christian Szegedy, Wei Liu, Yangqing Jia, Pierre Sermanet, Scott Reed, Dragomir
  Anguelov, Dumitru Erhan, Vincent Vanhoucke, and Andrew Rabinovich.
\newblock Going deeper with convolutions.
\newblock In \emph{Computer vision and pattern recognition}, pages 1--9, 2015.

\bibitem[Szegedy et~al.(2017)Szegedy, Ioffe, Vanhoucke, and
  Alemi]{szegedy2017inception}
Christian Szegedy, Sergey Ioffe, Vincent Vanhoucke, and Alexander~A Alemi.
\newblock Inception-v4, inception-resnet and the impact of residual connections
  on learning.
\newblock In \emph{AAAI}, volume~4, page~12, 2017.

\bibitem[Ulyanov et~al.(2016)Ulyanov, Vedaldi, and Lempitsky]{UlyanovVL16}
Dmitry Ulyanov, Andrea Vedaldi, and Victor Lempitsky.
\newblock Instance normalization: The missing ingredient for fast stylization.
\newblock \emph{arXiv preprint arXiv:1607.08022}, 2016.

\bibitem[Wu and He(2018)]{groupnorm}
Yuxin Wu and Kaiming He.
\newblock Group normalization.
\newblock In \emph{Computer Vision and Pattern Recognition}, 2018.

\end{thebibliography}
}

\end{document}